\documentclass{article}

\usepackage{microtype}
\usepackage{graphicx}
\usepackage{subfigure}
\usepackage{booktabs} 
\usepackage[dvipsnames]{xcolor}

\usepackage{hyperref}

\usepackage[accepted]{mlsys2023}

\usepackage{amsmath,amsfonts,bm}

\def\eqref#1{equation~\ref{#1}}

\def\1{\bm{1}}

\def\vtheta{{\bm{\theta}}}
\def\va{{\bm{a}}}
\def\vb{{\bm{b}}}

\def\ve{{\bm{e}}}

\def\vg{{\bm{g}}}

\def\vs{{\bm{s}}}
\def\vt{{\bm{t}}}

\def\vx{{\bm{x}}}
\def\vy{{\bm{y}}}

\def\mA{{\bm{A}}}
\def\mB{{\bm{B}}}
\def\mC{{\bm{C}}}

\def\mF{{\bm{F}}}
\def\mG{{\bm{G}}}
\def\mH{{\bm{H}}}
\def\mI{{\bm{I}}}
\def\mJ{{\bm{J}}}

\def\mP{{\bm{P}}}

\DeclareMathAlphabet{\mathsfit}{\encodingdefault}{\sfdefault}{m}{sl}
\SetMathAlphabet{\mathsfit}{bold}{\encodingdefault}{\sfdefault}{bx}{n}

\def\gB{{\mathcal{B}}}

\def\gL{{\mathcal{L}}}

\def\gO{{\mathcal{O}}}

\def\gT{{\mathcal{T}}}

\def\gX{{\mathcal{X}}}

\newcommand{\E}{\mathbb{E}}

\newcommand{\R}{\mathbb{R}}

\usepackage{booktabs,multirow}
\usepackage{tablefootnote}
\usepackage{enumitem} 
\usepackage{amssymb}
\usepackage{multirow}
\usepackage{url}

\usepackage{pifont}
\newcommand{\xmark}{\ding{55}}%

\newcommand{\sharpness}{sharpness}%
\newcommand{\gcov}{grad cov}%
\newcommand{\gsecm}{grad 2$^{\rm nd}$m}%
\newcommand{\gcovsecm}{grad cov, 2$^{\rm nd}$m}%

\definecolor{matpltblue}{HTML}{1f77b4}
\definecolor{matpltorange}{HTML}{ff7f0e}
\definecolor{matpltgreen}{HTML}{2ca02c}
\definecolor{matpltpurple}{HTML}{9467bd}
\definecolor{matpltyellow}{HTML}{bcbd22}
\definecolor{matpltred}{HTML}{d62728}
\definecolor{matpltbrown}{HTML}{8c564b}

\newcommand{\T}{\top}
\newcommand{\params}{\vtheta}

\newcommand{\inputs}{\vx}
\newcommand{\targets}{\vt}

\newcommand{\nn}{f}

\newcommand{\minibatch}{\gB}

\newcommand{\loss}{\gL}
\newcommand{\peloss}{\ell}

\newcommand{\avg}[1]{\left\langle#1\right\rangle}

\newcommand{\nparams}{P}

\newcommand{\nclasses}{K}

\newcommand{\damp}{\tau}

\definecolor{ForestGreen}{RGB}{34,139,34}
\definecolor{RoyalPurple}{HTML}{613F99}
\definecolor{Violet}{HTML}{58429B}
\definecolor{BlueViolet}{HTML}{473992}
\definecolor{Plum}{HTML}{92268F}

\newcommand{\hess}{{\color{matpltred}\mH}}
\newcommand{\hessbfgs}{\color{matpltbrown}\hat{\mH}_{\mathrm{bfgs}}}
\newcommand{\fim}{{\color{matpltorange}\mF}}
\newcommand{\fimnmc}{{\color{matpltgreen}\hat{\mF}_{\nmc{\rm mc}}}}
\newcommand{\fimmc}{{\color{matpltgreen}\hat{\mF}_{1{\rm mc}}}}
\newcommand{\fimemp}{{\color{matpltblue}\hat{\mF}_{\rm emp}}}
\newcommand{\fimempbatch}{{\color{matpltpurple}\hat{\mF}_{\rm emp}^{\rm batch}}}
\newcommand{\ggn}{{\color{orange}\mG}}

\newcommand{\sqrtfimempbatch}{{(\fimempbatch)^{1/2}}}

\newcommand{\hessabs}{\hess_{|\lambda|}}

\newcommand{\nmc}{n}

\newcommand{\metric}{\mC}

\usepackage{xcolor}

\mlsystitlerunning{ASDL: A Unified Interface for Gradient Preconditioning in PyTorch}

\begin{document}

\twocolumn[
\mlsystitle{ASDL: A Unified Interface for Gradient Preconditioning in PyTorch}




\begin{mlsysauthorlist}
\mlsysauthor{Kazuki Osawa}{eth}
\mlsysauthor{Satoki Ishikawa}{tt}
\mlsysauthor{Rio Yokota}{tt}
\mlsysauthor{Shigang Li}{eth}
\mlsysauthor{Torsten Hoefler}{eth}
\end{mlsysauthorlist}

\mlsysaffiliation{eth}{Department of Computer Science, ETH Zurich, Switzerland}
\mlsysaffiliation{tt}{Tokyo Institute of Technology, Japan}

\mlsyscorrespondingauthor{Kazuki Osawa}{kazuki.osawa@inf.ethz.ch}

\mlsyskeywords{Machine Learning, MLSys}

\vskip 0.3in

\begin{abstract}
\textit{Gradient preconditioning} is a key technique to integrate the \textit{second-order information} into gradients for improving and extending gradient-based learning algorithms.
In deep learning, stochasticity, nonconvexity, and high dimensionality lead to a wide variety of gradient preconditioning methods, with implementation complexity and inconsistent performance and feasibility.
We propose the Automatic \textit{Second-order} Differentiation Library (ASDL), an extension library for PyTorch, which offers various implementations and a plug-and-play \textit{unified interface} for gradient preconditioning. 
ASDL enables the study and structured comparison of a range of gradient preconditioning methods.
\end{abstract}
]



\printAffiliationsAndNotice{}  

\section{Introduction}
\textit{Gradient preconditioning} is a key technique for integrating  \textit{second-order} information such as \textit{loss sharpness} (second-order derivatives) and \textit{gradient covariance/second moment} (second-order statistics) into gradients.
In deep learning in various domains such as vision \citep{osawa_large-scale_2019}, language \cite{anil_scalable_2021,pauloski_deep_2022}, graph \citep{izadi_optimization_2020}, reinforcement learning \cite{kakade_natural_2002}, and quantum computing \citep{stokes_quantum_2020}, gradient preconditioning has been reported to \textit{improve} and \textit{extend} gradient-based learning algorithms.
The benefits of gradient preconditioning include 
faster convergence of training \citep{amari_natural_1998,martens_optimizing_2015},
more robust approximate Bayesian inference \citep{khan_fast_2018,zhang_noisy_2018,nado_stochastic_2018}, 
regularization to avoid forgetting in continual learning \citep{kirkpatrick_overcoming_2017,pan_continual_2020}, 
identifying influential parameters and examples on model's output \citep{hassibi_second_1993,koh_understanding_2017}, 
estimation of the mini-batch size with high data efficiency \citep{mccandlish_empirical_2018}, 
and generic probabilistic prediction via gradient boosting \citep{duan_ngboost_2020}. 

\begin{figure*}
    \centering
    \includegraphics[width=.95\textwidth]{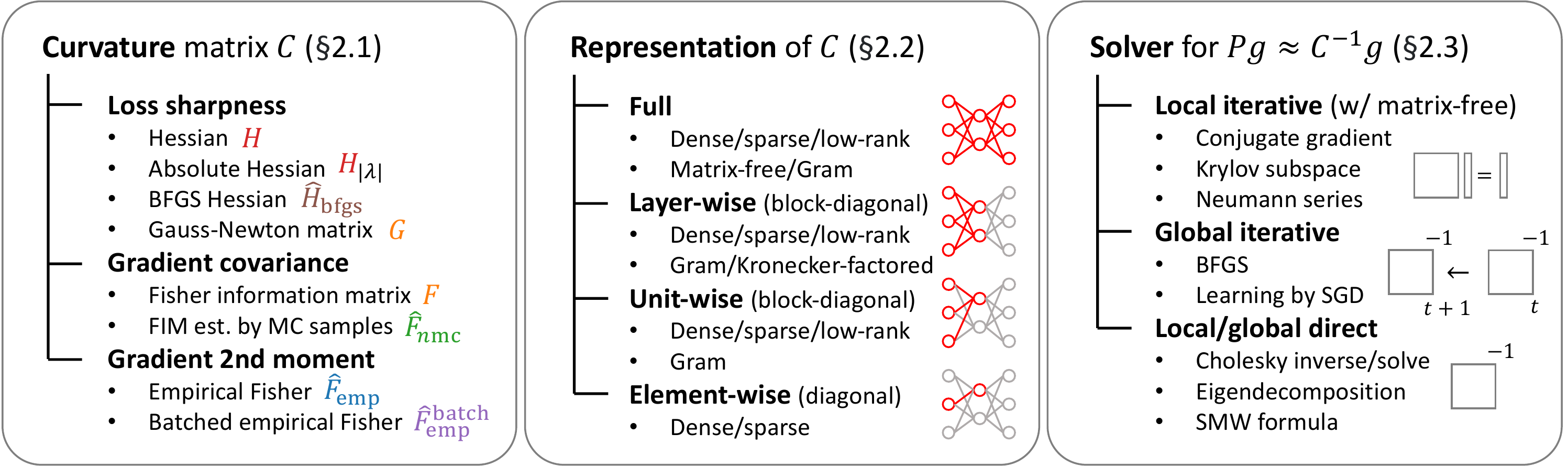}
    \caption{Three key components of gradient preconditioning in deep learning 
    }
    \label{fig:prec}
\end{figure*}

To integrate the second-order information into the gradient $\vg$,
the gradient preconditioning applies the \textit{preconditioning matrix} $\mP$ to get the \textbf{preconditioned gradient} $\mP\vg$.
In deep learning, where stochasticity, nonconvexity, and high dimensionality are inherent, there are a variety of choices for (i) the \textit{curvature matrices} $\metric$ containing various forms of second-order information (\S\ref{subsec:curv}), (ii) the \textit{representations} of $\metric$ based on the neural network structures and matrix properties (\S\ref{subsec:repr}), and (iii) the \textit{solvers} for computing $\mP\vg\approx\metric^{-1}\vg$ (\S\ref{subsec:solver}).
This leads to a \textit{diverse set} of gradient preconditioning methods (\autoref{fig:prec},\autoref{tab:representative}), each requiring \textit{algorithm-specific} and \textit{complex} implementations, making it challenging to incorporate them into existing training pipelines that usually use SGD-based gradient methods today. Furthermore, it is hard to switch between different methods in order to compare them.
This implementation issue is critical because the \textit{compute performance}, \textit{prediction accuracy}, and \textit{feasibility} (in terms of budget of time and memory) of methods are \textit{highly dependent} on neural network architectures and specific training settings (\S\ref{sec:case}).

To address this, we propose the Automatic \textit{Second-order} Differentiation Library (\textbf{ASDL}), which extends PyTorch \citep{paszke_pytorch_2019}, an automatic-differentiation library, with a \textbf{unified interface} for gradient preconditioning using various curvature matrices, representations, and solvers (\S\ref{subsec:interface}, \autoref{fig:interface}) that is compatible with several types of training pipelines and neural network architectures (\S\ref{subsec:versatile}).
ASDL has a hierarchical abstraction structure (\S\ref{subsec:structure}) that facilitates the development and optimization of various gradient preconditioning methods.
We use ASDL to apply gradient preconditioning methods for optimization (i.e., second-order optimization and adaptive gradient methods) to mini-batch gradient-based training of MLPs, CNNs, and Transformers.
We observe the throughput (example/s), peak memory consumption, and generalization performance with varying the neural network architecture, hyperparameters (e.g., mini-batch size, matrix update interval), and gradient preconditioning method and discuss an intriguing relationship between them (\S\ref{sec:case}).

\newpage

\begin{table*}
    \centering
    \caption{
    {Representative gradient preconditioning methods in deep learning}. 
    ``\textbf{KF}'': Kronecker-factored. 
    ``\textbf{RR}'': Rank reduction. 
    ``\textbf{SMW}'': Sherman-Morrison-Woodbury formula.
    Methods analyzed in this study are \underline{underlined}. 
    See \autoref{tab:prec_full} for a more comprehensive list. 
    }
    \resizebox{\textwidth}{!}{
        \begin{tabular}{lllllll}
        \toprule
        \multirow{2}[2]{*}{{Method}} & \multicolumn{2}{c}{\textbf{Curvature} matrix $\metric$ (\S\ref{subsec:curv})} & \multicolumn{2}{c}{\textbf{Representation} of $\metric$ (\S\ref{subsec:repr})} & \multicolumn{2}{c}{\textbf{Solver} for  $\mP\vg\approx\metric^{-1}\vg$ (\S\ref{subsec:solver})} \\
        \cmidrule(lr){2-3} \cmidrule(lr){4-5} \cmidrule(lr){6-7}
        & type & matrix & granularity & format & type & key operations \\
        \midrule
        Hessian-free \citep{martens_deep_2010} & \sharpness & $\hess,\ggn$ & full & matrix-free & local iterative & conjugate gradient \\
        \underline{PSGD (KF)} \citep{li_preconditioned_2018} & \sharpness & $\hessabs$ & layer & KF & global iterative  & triangular solve, SGD \\
        \underline{K-BFGS} \citep{goldfarb_practical_2021} & \sharpness & $\hessbfgs$ & layer & KF & global iterative  & BFGS \\
        \underline{K-FAC} \citep{martens_optimizing_2015} & \gcovsecm & $\fimmc,\fimemp$ & layer & KF & local/global direct & Cholesky inverse \\
        \underline{SENG} \citep{yang_sketch-based_2022} & \gsecm & $\fimemp$ & layer & Gram, RR & local direct & SMW inverse, sketching \\
        \underline{Shampoo} \citep{gupta_shampoo_2018} & \gsecm & $\sqrtfimempbatch$ & layer & KF & global direct & eigendecomp. \\
        Adam \citep{kingma_adam_2015} & \gsecm & $\sqrtfimempbatch$ & element & dense & global direct & element-wise division \\
        \bottomrule
        \end{tabular}
    }
    \label{tab:representative}
\end{table*}

\section{Gradient Preconditioning in Deep Learning}
\label{sec:taxonomy}

\paragraph{Notations}
The \textit{mini-batch empirical loss}
\begin{equation}
\label{eq:loss}
\begin{split}
\loss(\params)
&:=\frac{1}{|\minibatch|}\sum_{(\inputs,\targets)\in\minibatch}\peloss(\inputs,\targets;\params)
\\
&=\avg{\peloss(\inputs,\targets;\params)}
=\avg{h(\nn(\inputs),\targets)}
\end{split}
\end{equation}
is the average of the per-example negative log-likelihood $\peloss(\inputs,\targets;\params):=-\log p_\params(\targets|\inputs)=:h(\nn(\inputs;\params),\targets)$ for each input-target pair $(\inputs,\targets)$ ($\inputs\in\gX,\targets\in\gT$) in a mini-batch $\minibatch$ sampled from the training set.
$\params\in\R^{\nparams}$ is the column vector containing the neural network parameters, $\avg{\cdot}$ represents the average over $\minibatch$, $p_\params$ is model's predictive distribution, $q$ is input distribution, $\nn:\gX\to\R^{\nclasses}$ is the neural network with $\nclasses$ output neurons parameterized by $\params$,
$h:\R^{\nclasses}\times\gT\to\R$ evaluates the negative log-likelihood for output-target pair, 
$\vg:=\nabla\loss(\params)\in\R^{\nparams}$ is the \textit{mini-batch gradient}, and $\mJ_\nn(\inputs)\in\R^{\nclasses\times\nparams}$ is the Jacobian of $\nn$ with respect to (w.r.t.) $\params$.

\subsection{Curvature matrices}
\label{subsec:curv}
\paragraph{Loss sharpness}
The \textit{Hessian} matrix    
\begin{align}
\label{eq:hess}
\hess:=\nabla^2\loss=\avg{\nabla^2\peloss(\inputs,\targets;\params)}\in\R^{\nparams\times\nparams}
\end{align}
is the second-order derivative of $\loss$ representing the \textit{loss sharpness} \citep{hochreiter_flat_1997}, and the Newton direction is $\mP\vg=\hess^{-1}\vg$  .
The \textit{absolute Hessian} $\hessabs$, which replaces the eigenvalues of $\hess$ by their absolute values, is preferred in optimization of a nonconvex $\loss$ to avoid saddle points \citep{dauphin_identifying_2014,li_preconditioned_2018} 
and $\mP=\hessabs^{-1}$ is the only positive definite matrix that perfectly reduces (i.e., to 1) the condition number of $\mP\hess$ \citep{dauphin_equilibrated_2015}.
The BFGS method estimates $\hess$ (or $\hess^{-1}$) with the \textit{BFGS Hessian} $\hessbfgs$ (or ${\hessbfgs}^{-1}$), which is the accumulation of the changes in $\vg$ (i.e., changes in the first-order derivatives) and $\params$ during iterative optimization of $\params$ with $\mP\vg={\hessbfgs}^{-1}\vg$:
\begin{align}
\label{eq:bfgs}
   \mB_{t+1}
   \leftarrow
   \mB_t
   +
   \frac{\vy_t\vy_t^\T}{\vy_t^\T\vs_t}
   -
   \frac{\mB_t\vs_t\vs_t^\T\mB_t^\T}{\vs_t^\T\mB_t\vs_t}\,,
\end{align}
where $\mB_t\in\R^{\nparams\times\nparams}$ is $\hessbfgs$, $\vy_t=\vg_{t+1}-\vg_t\in\R^\nparams$, and $\vs_t=\mP_t\vg_t\in\R^\nparams$ at $t$-th optimization step.
The (generalized) \textit{Gauss-Newton matrix} \cite{schraudolph_fast_2002}
\begin{align}
\label{eq:ggn}
\ggn:=\avg{\mJ_\nn(\vx)^\T\nabla_\vy^2h(\vy,\targets) |_{\vy=\nn(\inputs)}\mJ_\nn(\vx)}
\end{align}
, which ignores the second-order derivative of $\nn$ w.r.t. $\params$ in $\hess$ (i.e., views $\nn$ as linear \cite{grosse_chapter_2022}) and is positive semi-definite, is also preferred in non-convex optimization \citep{martens_deep_2010}. 

\paragraph{Gradient covariance}
The \textit{Fisher information matrix}
\begin{align}
\label{eq:fim}
\fim:=\E_{q(\inputs)}\left[\E_{p_\params(\targets'|\inputs)}\left[\nabla\log p_\params(\targets'|\inputs)\nabla\log p_\params(\targets'|\inputs)^\T\right]\right]
\end{align}
$\in\R^{\nparams\times\nparams}$ is the \textit{covariance} of gradient of log-likelihood $\nabla\log p_\params$.
$\fim$ is also the second-order derivative of the KL-divergence $D_{\rm KL}(p_{\params}||p_{\params+\Delta\params})$ and is used as $\metric$ in the natural gradient descent (NGD) \citep{amari_natural_1998}: $\mP\vg=\fim^{-1}\vg$. 
In practice, $\E_{q(\inputs)}[\cdot]$ is estimated with $\avg{\cdot}$, and $\fim=\ggn$ for cross-entropy and MSE loss \cite{pascanu_revisiting_2014}, connecting the loss sharpness and gradient covariance perspectives in optimization \citep{martens_new_2020}.
$\E_{p_\params(\targets'|\vx)}[\cdot]$ involves $\nclasses$ backward passes for $\nabla\log p_\params$ \citep{dangel_backpack_2020} (e.g., $\nclasses=1000$ for ImageNet-1K), 
so $\fim$ is often estimated with the \textit{MC Fisher} with $n$ Monte-Carlo (MC) samples of $\targets_{\rm mc}\sim p_\params(\targets'|\inputs)$:
\begin{align}
\label{eq:fimnmc}
\fimnmc:=\avg{\sum_{i=1}^n\nabla\log p_\params(\targets_{\rm mc}^{(i)}|\inputs)\nabla\log p_\params(\targets_{\rm mc}^{(i)}|\inputs)^\T}
\end{align}
$n=1$, i.e., $\fimmc$, is often used \citep{martens_optimizing_2015}.

\paragraph{Gradient second moment}
The \textit{empirical Fisher} 
\begin{equation}
\label{eq:fimemp}
\begin{split}
\fimemp&
:=\avg{\nabla\ell(\inputs,\targets;\params)\nabla\ell(\inputs,\targets;\params)^\T}
\\
&=\avg{\nabla\log p_\params(\targets|\inputs)\nabla\log p_\params(\targets|\inputs)^\T}
\in\R^{\nparams\times\nparams}
\end{split}
\end{equation}
is the \textit{second moment} of \textit{per-example} empirical gradient. It can be computed during the backward pass for the \textit{empirical} gradient $\nabla\loss$ and is preferred in large-scale settings \citep{osawa_large-scale_2019,pauloski_deep_2022}.
As $\fimemp$ is no longer centered (i.e., $\avg{\nabla\peloss(\inputs,\targets;\params)}\ne\mathbf{0}$), it is claimed not to capture the useful second-order information for optimization \citep{kunstner_limitations_2020} while it is empirically observed that NGD with $\fimemp$ still achieves the fast convergence with smoothed $\targets$ \citep{pauloski_deep_2022,osawa_scalable_2022}.
Adaptive gradient methods such as Adam \cite{kingma_adam_2015} and Shampoo \citep{gupta_shampoo_2018} use the \textit{batched empirical Fisher} 
\begin{align}
\label{eq:fimempbatch}
\fimempbatch(T):=\sum_{t=1}^T\alpha_t\vg_t\vg_t^\T
\,\,\,
(0\leq\alpha_t\leq 1)
\end{align}
where $\vg_t$ is for $\minibatch_t$ at $t$-th training step, an online estimate of the \textit{second moment} of \textit{mini-batch} empirical gradient: $\mP\vg_T=(\fimempbatch(T))^{-1/2}\vg_T$.
$\fimempbatch$ looses the second-order information when the mini-batch size $|\minibatch|$ is large \citep{grosse_chapter_2022}, but it is also empirically observed that Shampoo achieves a faster convergence than first-order optimizers (SGD, LAMB \citep{you_large_2017}) in large-batch training \citep{anil_scalable_2021}\footnote{See \cite{grosse_chapter_2022} for a more detailed description of these curvature matrices.}.

\subsection{Representations of matrices}
\label{subsec:repr}
It is infeasible to materialize $ \metric\in\R^{\nparams\times\nparams}$ 
and directly invert it , i.e., $\metric^{-1}$, with the $\gO(\nparams^3)$ cost 
for deep neural networks with a massive number of parameters $\nparams$, e.g., billions. 
To make practical use of (a portion of) the information in $\metric$, there are various \textit{matrix representations} using compact format, block-diagonal approximation, or both.
\paragraph{Full matrix}
Typical compact formats for exploiting the \textit{full} $\metric$ include matrix-vector products (\textit{matrix-free}), e.g., Hessian-free \citep{martens_deep_2010}, and \textit{Gram matrix} with rank reduction, e.g., SMW-NG \citep{ren_efficient_2019}.
\paragraph{Layer-/unit-/element-wise block-diagonal matrix}
Granularity of diagonal blocks are often per neural network \textit{layer}, per \textit{unit}, or per \textit{element} of $\params$ (i.e., diagonal, e.g., Adam). 
Layer-wise blocks are still too large to be materialized in most of today's deep neural network architectures, e.g., Transformers \citep{vaswani_attention_2017}. 
For layer-wise blocks, one of the most common compact formats is \textit{Kronecker-factored matrix}, where each layer-wise block is approximated with the Kronecker product of two (much smaller) matrices or more, e.g., PSGD \citep{li_preconditioned_2018}, K-BFGS \citep{goldfarb_practical_2021}, K-FAC \citep{martens_optimizing_2015}, Shampoo \citep{gupta_shampoo_2018}.

\subsection{Solvers for preconditioning gradient}
\label{subsec:solver}
\paragraph{Local vs. global}
\textit{Solvers} to get $\mP\vg\approx\metric^{-1}\vg$ are first classified by the scope of information captured by $\metric$, i.e., \textit{local} information within one $\minibatch$ observed at one time step vs. \textit{global} information associated with multiple $\minibatch$s observed through multiple time steps (with different models). 
By definition, solvers with $\hessbfgs$ or $\fimempbatch$ are global solvers.
\paragraph{Iterative vs. direct}
Solvers are also classified by the type of linear solver for $\metric_{repr}\vx=\vg$, i.e., \textit{iterative} vs. \textit{direct}, where $\metric_{repr}$ is a certain representation (\S\ref{subsec:repr}) of selected $\metric$ (\S\ref{subsec:curv}) containing local or global information.
An iterative local solver uses the matrix-free format while an iterative global solver materializes $\metric_{repr}$.
A damping $\damp\mI$ ($\damp>0$) is often added to $\metric_{repr}$ to improve numerical stability and/or guarantee positive definiteness ($\left(\metric_{repr}+\damp\mI\right)\succ 0$).
This allows a fast direct solver using Cholesky decomposition (e.g., K-FAC) or Sherman-Morrison-Woodbury (SMW) formula \cite{petersen_matrix_2012} (e.g., SMW-NG, SENG \citep{yang_sketch-based_2022}) to be applied.

\autoref{tab:representative} lists representative gradient preconditioning methods with a selection of different types of components.

\begin{figure*}
    \centering
    \includegraphics[width=.93\textwidth]{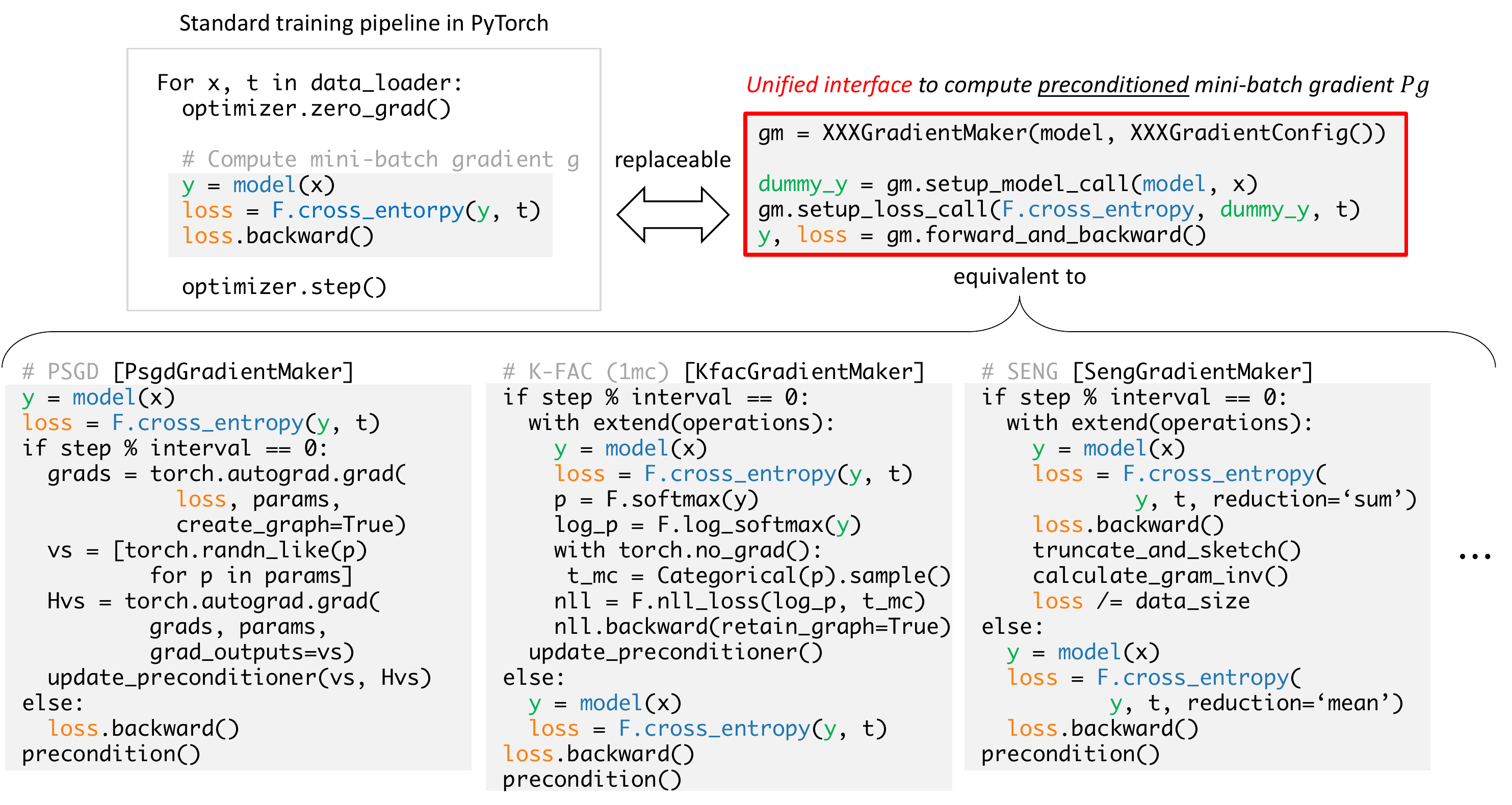}
    \caption{\textbf{Unified interface for gradient preconditioning in PyTorch.} 
    \texttt{XXXGradientMaker} (``\texttt{XXX}'': algorithm name), offered by ASDL, hides \textit{algorithm-specific} and \textit{complex} operations for $\mP\vg$ in a \textit{unified} way. 
    For training without gradient preconditioning, \texttt{GradientMaker} computes $\vg$ with \textit{the same interface} (i.e., no need to switch scripts).
    }
    \label{fig:interface}
\end{figure*}

\section{Automatic Second-order Differentiation Library (ASDL)}
Our Automatic \textit{Second-order} Differentiation Library (\textbf{ASDL})\footnote{\url{https://github.com/kazukiosawa/asdl}} implements
gradient preconditioning methods listed in \autoref{tab:representative} and (a large portion of) \autoref{tab:prec_full}. 
We now introduce the programming interface of ASDL  (\S\ref{subsec:interface}), its usage in various situations and its versatility (\S\ref{subsec:versatile}), and ASDL's code structure (\S\ref{subsec:structure}).

\subsection{Unified interface for gradient preconditioning}
\label{subsec:interface}
\autoref{fig:interface} shows a common training pipeline in PyTorch with mini-batch gradients $\vg$, the (simplified) operations in PSGD, K-FAC (with $\fimmc$), and SENG, 
and the \textbf{unified interface} in ASDL, \texttt{XXXGradientMaker} class (``\texttt{XXX}'': algorithm name), which enables an easy integration of gradient preconditioning $\mP\vg$ by hiding the \textit{algorithm-specific} and \textit{complex} operations.
The behavior of the gradient preconditioning is defined by the \texttt{XXXGradientMaker} class and is configured by the passed \texttt{XXXGradientConfig} object.
For example, to perform PSGD, K-FAC, or SENG, one can initialize \texttt{gm} in \autoref{fig:interface} with \texttt{PsgdGradientMaker}, \texttt{KfacGradientMaker}, or \texttt{SengGradientMaker}, respectively.
For convenience, ASDL also offers a \texttt{GradientMaker} class for calculating $\vg$ (without gradient preconditioning). 
To perform the (preconditioned) gradient calculation in a unified way, \texttt{XXXGradientMaker} and \texttt{GradientMaker} have the following \textit{common} APIs:
\begin{enumerate}
    \item \texttt{\textbf{setup\_model\_call(}model\_fn, *args, **kwargs\textbf{)}}: The first argument (\texttt{model\_fn}) is a function (typically an object of \texttt{torch.nn.Module}) that performs a forward pass on the neural network $\nn$ (and the loss function $h$, depending on the definition of \texttt{model\_fn}) and returns a certain format of the output, and \texttt{*args} and \texttt{**kwargs} are the arguments to \texttt{model\_fn}.
    This method returns a \texttt{\textbf{DummyObject}}, which behaves as if it were the actual output of \texttt{model\_fn} (which has not yet been evaluated at this point) and can be used to define how the loss value should be evaluated (examples in \autoref{subsec:versatile}).
    \item \texttt{\textbf{setup\_loss\_call(}loss\_fn, *args, **kwargs\textbf{)}}: The first argument (\texttt{loss\_fn}) is a function that evaluates the loss function $h$
    , and \texttt{*args} and \texttt{**kwargs} are the arguments to \texttt{loss\_fn}. The output of \texttt{model\_fn} (or its modification), i.e., \texttt{DummyObject}, can be an argument to \texttt{loss\_fn} (examples in \autoref{subsec:versatile}).   
    \item \texttt{\textbf{setup\_loss\_repr(}loss\_repr\textbf{)}}: An alternative of \texttt{setup\_loss\_call}. The argument (\texttt{loss\_repr}) is a \texttt{DummyObject} that specifies how the loss value should be represented based on the output of \texttt{model\_fn} (examples in \autoref{subsec:versatile}).
    \item \texttt{\textbf{forward\_and\_backward()}}: After setting up \texttt{model\_fn} and \texttt{loss\_fn} (or \texttt{loss\_repr}), this method performs a forward pass (by calling both with the specified arguments) and a backward pass on them to calculate $\vg$ or $\mP\vg$. The resulting (preconditioned) gradients are stored at \texttt{param.grad} (or accumulated to it if it exists) of each \texttt{param} (\texttt{torch.nn.parameter.Parameter}) of the \texttt{model} (\texttt{torch.nn.Module}) in the same way as \texttt{loss.backward()}.
    This method returns \texttt{model\_fn}'s output and loss value (either \texttt{loss\_fn}'s output or \texttt{loss\_repr}'s evaluation).
\end{enumerate}
As shown in \autoref{fig:interface} (and discussed in \autoref{subsec:versatile}), these procedures are \textit{algorithm-independent} and as \textit{simple} (same logical structure) as the standard training pipeline in PyTorch.
The unified interface in ASDL enables us to flexibly switch/compare methods, which is critical as each gradient preconditioning method exhibits compute performance, prediction accuracy, and feasibility depending \textit{highly} on neural network architectures and specific training settings (\autoref{sec:case}).  

\subsection{Versatility of the interface}
\label{subsec:versatile}
The idea behind the design of these APIs is to do only the ``setup'' outside and hide the \textit{evaluation} inside \texttt{forward\_and\_backward()}, since the proper timing/context of the model $\nn$ and loss $h$ evaluations depends on the gradient preconditioning method as described in \autoref{fig:interface}.
However, defining an interface in this way that is compatible with a wide range of training pipelines is not simple.
This is because (i) the format of the output of the \texttt{model\_fn} depends on the training pipeline, (ii) it is even possible that \texttt{model\_fn} includes both the model $\nn$ and loss $h$ evaluations, and (iii) the \texttt{loss\_fn} usually takes (a part of) the \textit{evaluated value} of \texttt{model\_fn} (or the result of manipulating it) as an argument, which we have to tell \texttt{forward\_and\_backward} before calling it, i.e., \textit{before evaluating} \texttt{model\_fn}.

\begin{figure}[t]
    \centering
    \includegraphics[width=\linewidth]{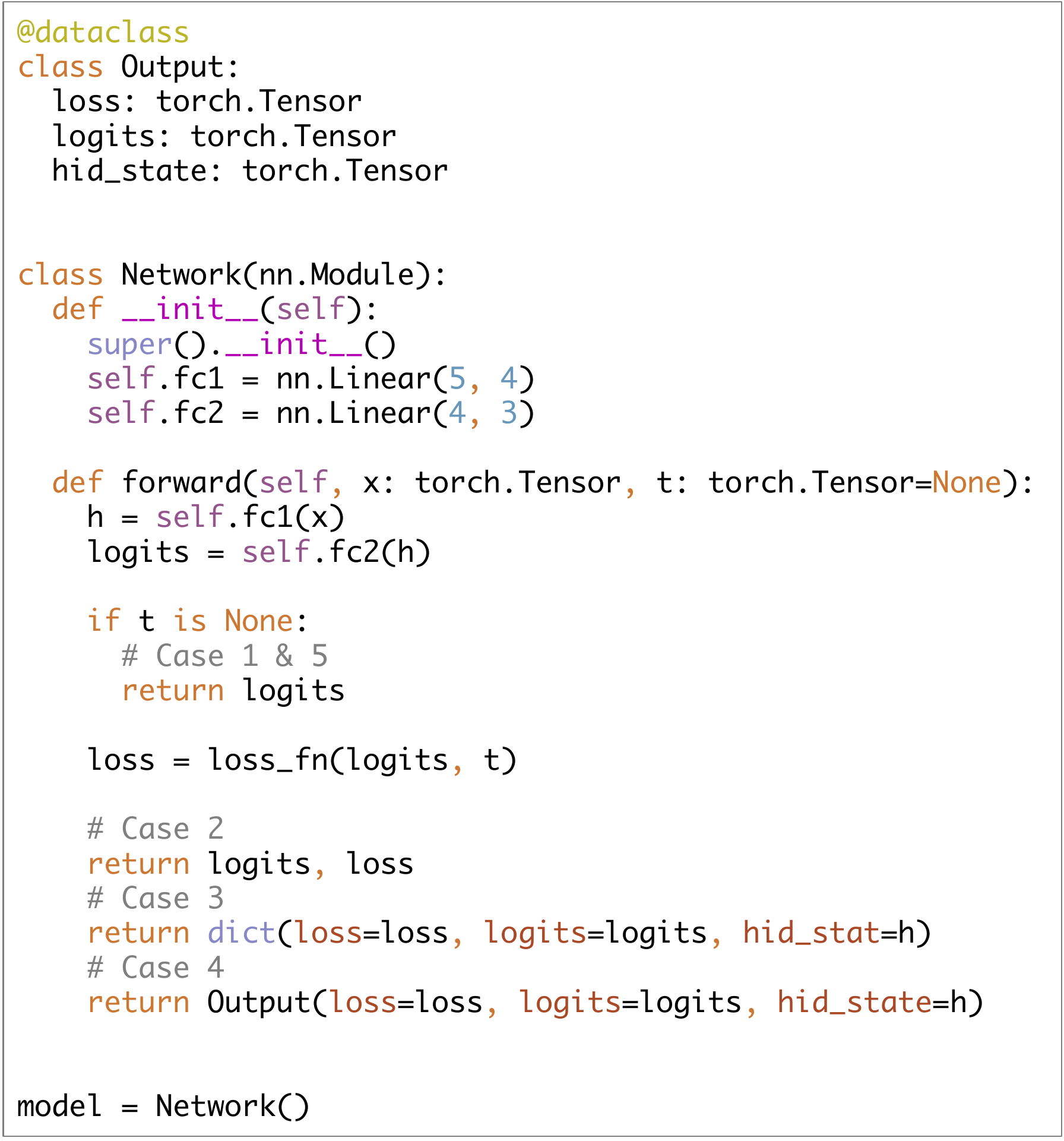}
    \caption{PyTorch \texttt{model} (an MLP) with different output formats}
    \label{fig:model}
\end{figure}

To address these challenges, \texttt{DummyObject} plays a key role in the APIs.
Below are some common training pipeline cases in PyTorch to demonstrate the versatility of the interface. 
For each case, we assume that the \texttt{model} is defined as a simple linear MLP with a certain output format as shown in \autoref{fig:model}.

\paragraph{Case 1: \texttt{torch.Tensor} output}
The first case is probably the most typical one, which is the same as what we consider in \autoref{fig:interface}.
The \texttt{model} receives an input \texttt{x} (\texttt{torch.Tensor}), which represents a batch of input examples (e.g., images) and returns the \texttt{y=logits} (\texttt{torch.Tensor}), which represents a batch of logits (a batch of $\nclasses$-dimensional vector).
The output \texttt{y} and the target \texttt{t} (\texttt{torch.Tensor}) are passed to \texttt{loss\_fn} to evaluate the loss value.
Finally, the mini-batch gradient $\vg$ is calculated by performing \texttt{loss.backward()}.
In ASDL, the same procedures can be written with a similar logical structure.
As we described in \autoref{subsec:interface}, \texttt{setup\_model\_call()} returns a \texttt{DummyObject} (\texttt{dum\_y} in the figure below).
\texttt{dum\_y} can be directly passed to \texttt{setup\_loss\_call()} in the same way that \texttt{y} is passed to \texttt{loss\_fn()}.
When \texttt{forward\_and\_backward} is called, \texttt{dum\_y} is replaced with the evaluated value and is passed to \texttt{loss\_fn()}, which is registered by \texttt{setup\_loss\_call()}.

\begin{figure}[h]
    \centering
    \frame{\includegraphics[width=\linewidth]{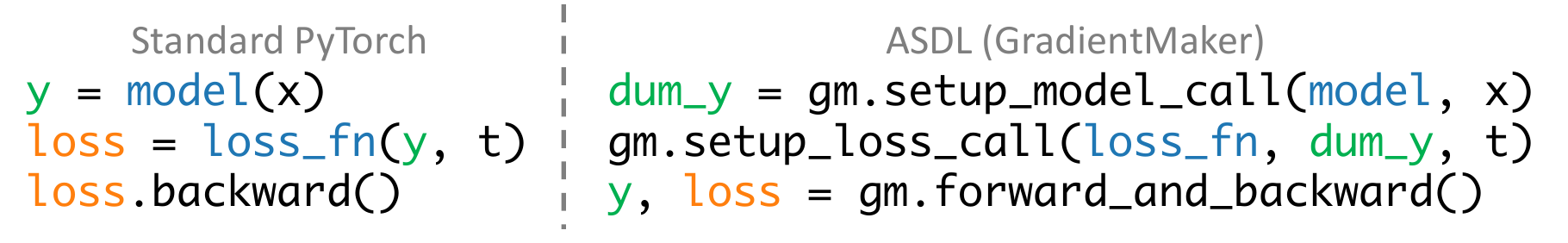}}
    \label{fig:case1}
\end{figure}

\paragraph{Case 2: Sequence (e.g., tuple, list) output}
Next, we consider the case where the loss evaluation is included in the \texttt{model} and it returns a tuple \texttt{(logits,loss)}.
Note that both input \texttt{x} and target \texttt{t} are passed to the \texttt{model} this time.
In this case, instead of \texttt{setup\_loss\_call}, we call \texttt{setup\_loss\_repr} to let the \texttt{GradientMaker} know how the loss value should be evaluated.
\texttt{dum\_y} behaves as if it were the actual value (tuple) and we know that the loss value would be stored in the second element of the tuple, so we can specify \texttt{dum\_y[1]} as the argument of \texttt{setup\_loss\_repr}. 

\begin{figure}[h]
    \centering
    \frame{\includegraphics[width=\linewidth]{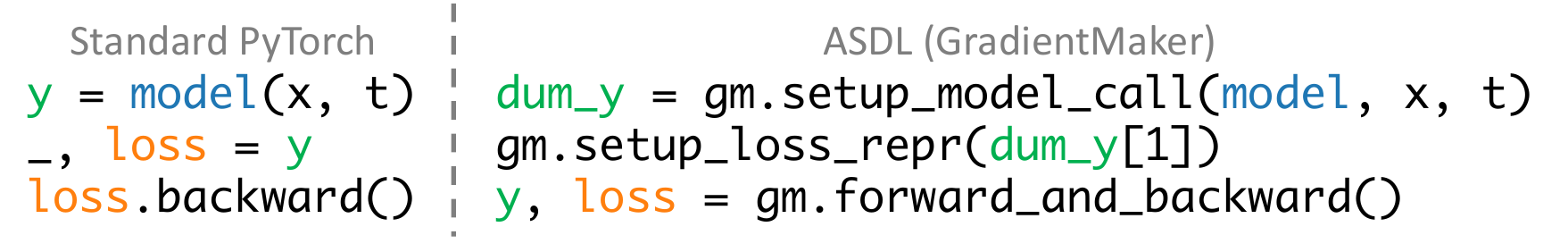}}
    \label{fig:case2}
\end{figure}

\paragraph{Case 3: Mapping (e.g, dict) output}
Similarly, the case where output \texttt{y} is a dictionary (or an arbitrary mapping object) is also supported in ASDL.

\begin{figure}[h]
    \centering
    \frame{\includegraphics[width=\linewidth]{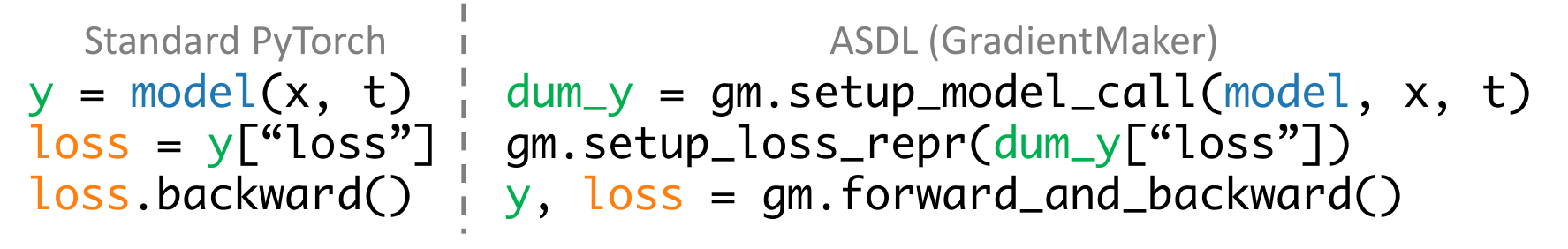}}
    \label{fig:case3}
\end{figure}

\paragraph{Case 4: \texttt{dataclass} output}
It is also common for the output \texttt{y} to be an object of the Python \texttt{dataclass}\footnote{\url{https://docs.python.org/3/library/dataclasses.html}} (or a some class for storing data). 
This case can be seen, for example, Hugginface's Transformers \cite{wolf_huggingfaces_2020}.
We can \textit{pseudo-access} the \texttt{loss} attribute (or an arbitrary attribute) through \texttt{dum\_y.loss} (or \texttt{dum\_y.attr\_name}).

\begin{figure}[h]
    \centering
    \frame{\includegraphics[width=\linewidth]{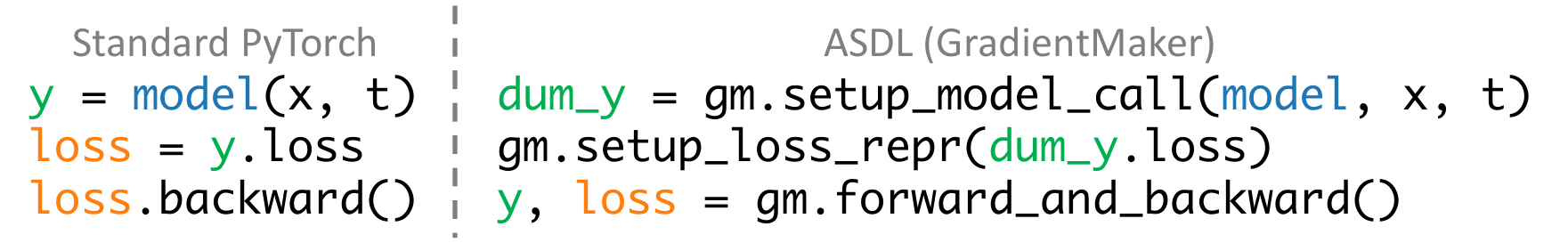}}
    \label{fig:case4}
\end{figure}

\paragraph{Case 5: Complex operations on output}
Finally, we consider the case in a language modeling task, where the input \texttt{x} is a \texttt{torch.Tensor} of shape \texttt{(batch\_size, sequence\_length, embedding\_dimension)} while the target \texttt{t} is a \texttt{torch.Tensor} of shape \texttt{(batch\_size, sequence\_length)} containing word ids in the vocabulary.
Here, the output \texttt{y} of the \texttt{model} have the shape \texttt{(batch\_size, sequence\_length, embedding\_dimension)}, and we wish to flatten \texttt{y} along the \texttt{batch\_size} and \texttt{sequence\_length} dimensions before evaluating the cross-entropy loss (\texttt{F.cross\_entropy}) by \texttt{y.view(-1, y.size(-1))}.
In ASDL, these operations can be expressed in the same way, i.e., \texttt{dum\_y.view(-1, dum\_y.size(-1))}.
It is possible to not only pseudo-access the attribute of \texttt{dum\_y} (e.g., \texttt{.view}), but also to \textit{pseudo-call} it (e.g., \texttt{.view()}).
Furthermore, we can pass \texttt{dum\_y} itself or the result of the pseudo-call to the pseudo-call.
Note once again that \texttt{dum\_y} does not contain the actual evaluation value at this point.
How can the \texttt{GradientMaker} know the actual \texttt{size} of \texttt{y} before evaluating it?
When \texttt{forward\_and\_backward} is called, the \texttt{GradientMaker} evaluates the sequence of the operations on the \texttt{DummyObject} (if any) \textit{recursively}.
This enables as complex operations on the output as this example.

\begin{figure}[h]
    \centering
    \frame{\includegraphics[width=\linewidth]{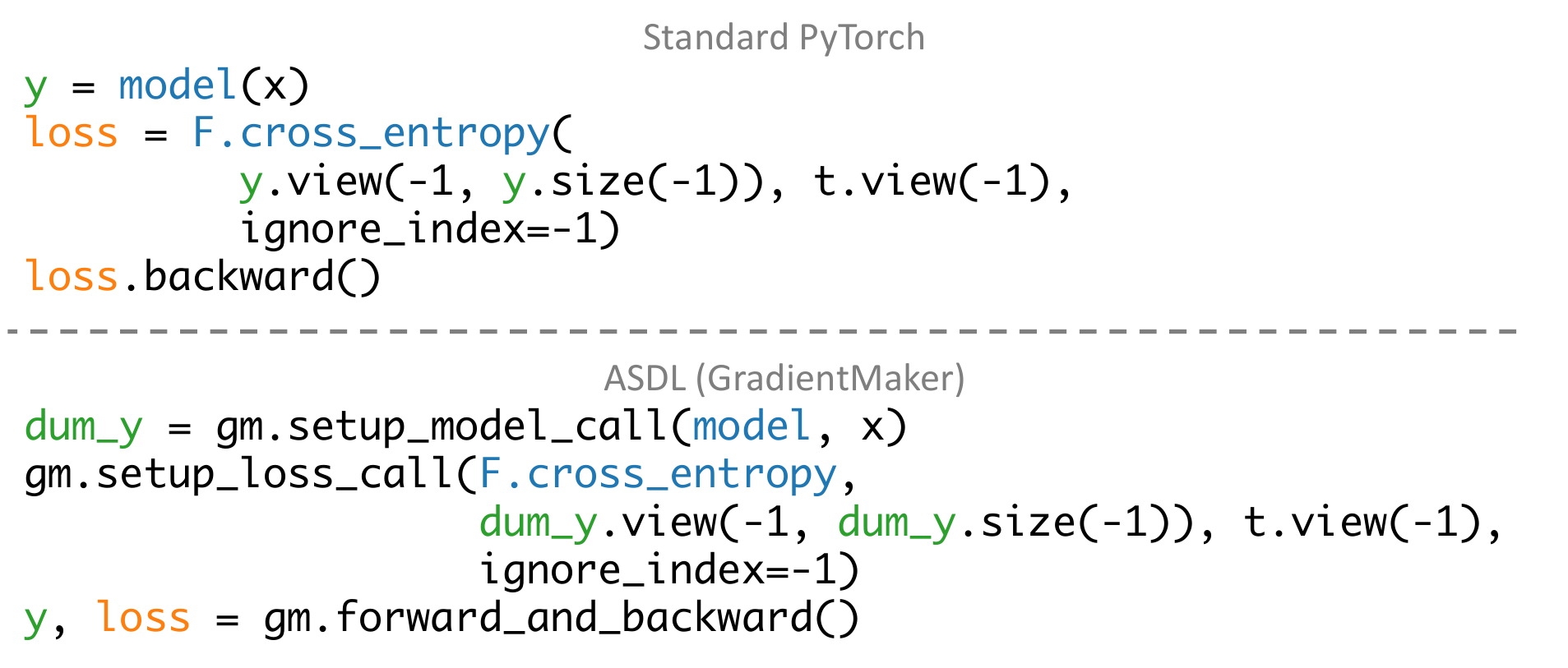}}
    \label{fig:case5}
\end{figure}

We have seen the versatility of ASDL's GradientMaker interface in five common cases. 
This flexibility is made possible by the expressive power of the \texttt{DummyObject} (\texttt{dum\_y)}.
Beyond the cases we have seen, one can manipulate a \texttt{DummyObject} with \textit{an arbitrary number of} \texttt{\_\_get\_item\_\_()}, \texttt{\_\_get\_attr\_\_()}, or \texttt{\_\_call\_\_()} operations in a recursive way, e.g., \texttt{dum\_y[0]["key"].attr.method(dum\_y[1])}.
When \texttt{forward\_and\_backward} is called, the series of operations are applied to the actual object (\texttt{y}) in exactly the same order. 
Therefore, it is the user's responsibility to ensure the validity of each operation, but that is also the case with standard PyTorch.
The flexibility provided by the \texttt{DummyObject} and the loss definition (\texttt{setup\_loss\_call} or \texttt{setup\_loss\_repr}) allows \texttt{XXXGradientMaker}, i.e., gradient preconditioning, to be integrated into a wide range of training pipelines in PyTorch with minimal development cost.

\subsection{Hierarchical structure of ASDL}
\label{subsec:structure}
ASDL supports various gradient preconditioning methods, which consist of different operations (e.g., automatic differentiation, matrix multiplication, matrix decomposition, and matrix inversion), depending on their components, i.e., curvature matrix (\S\ref{subsec:curv}), matrix representation (\S\ref{subsec:repr}), and solver (\S\ref{subsec:solver}). 
Furthermore, the definition of such operations can depend on the layer types (\texttt{torch.nn.Module}) that constitute the neural network.
To increase code reusability, maintainability, and extensibility, ASDL has a hierarchical abstraction structure, allowing for structured development and optimization of the implementations of various gradient preconditioning methods.

\begin{figure}[t]
    \centering
    \includegraphics[width=\linewidth]{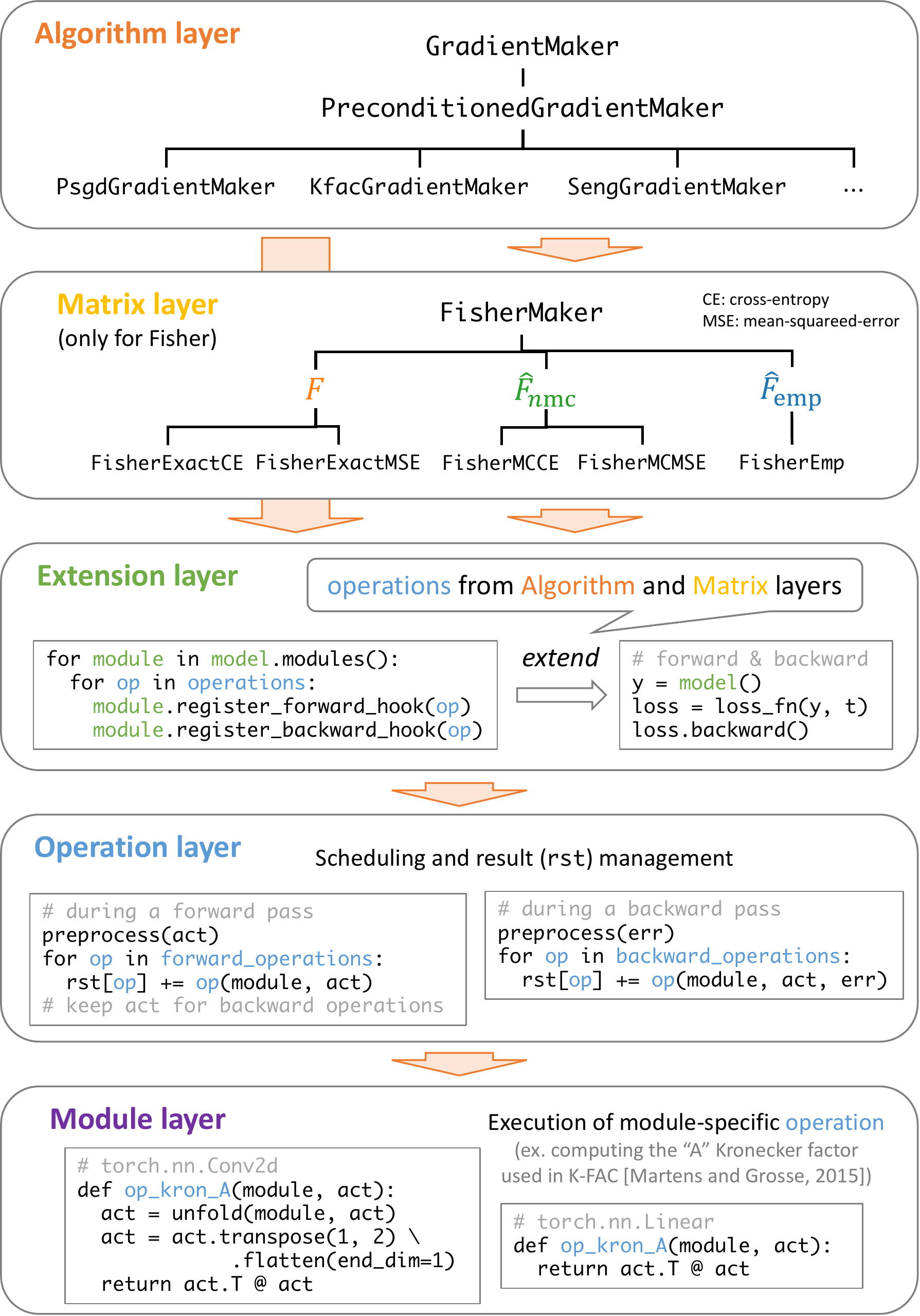}
    \caption{Abstraction layers in ASDL}
    \label{fig:layer}
\end{figure}

ASDL consists of five abstraction layers (\autoref{fig:layer}).

\begin{figure*}[t]
    \centering
    \includegraphics[width=\textwidth]{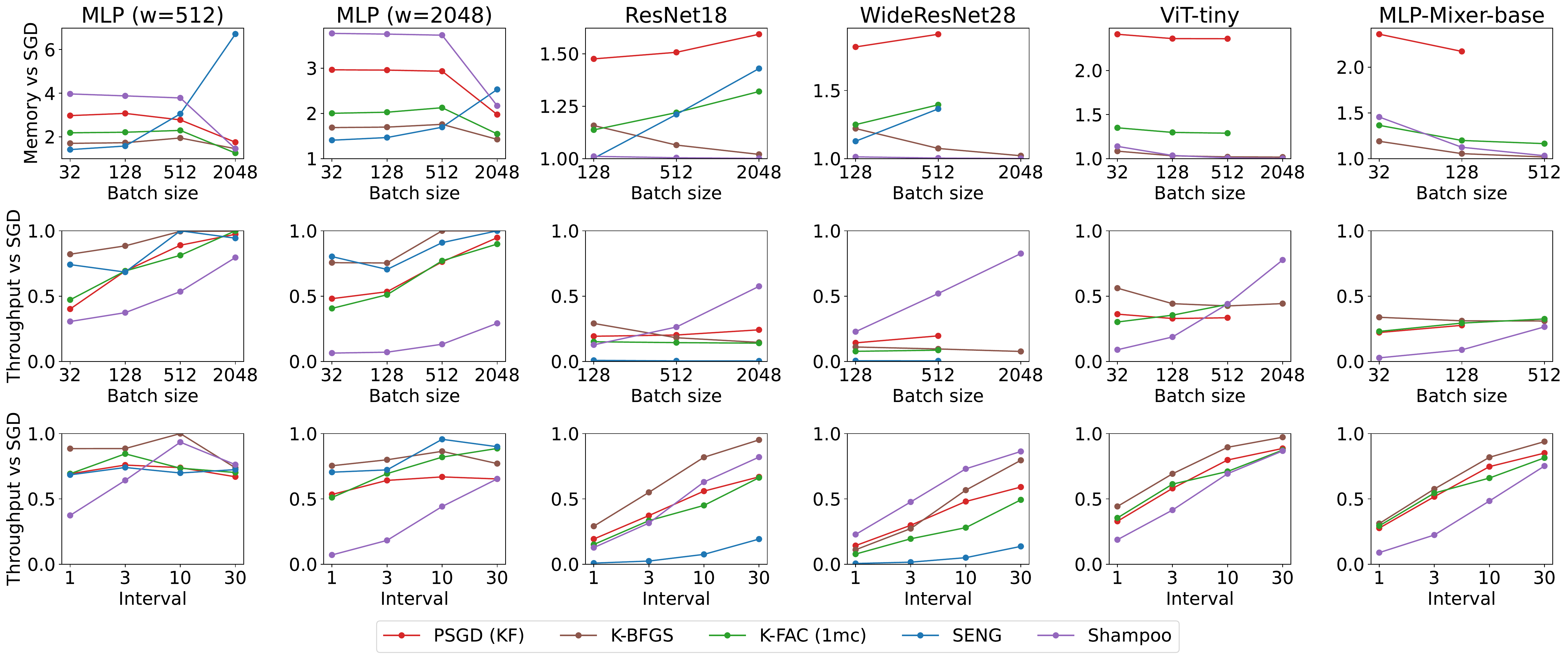}
    \caption{The ratio of peak memory ($\geq 1$) (top) and throughput [image/s] ($\leq 1$) (middle, bottom) of gradient preconditioning methods compared to SGD with various mini-batch sizes $|\minibatch|$ and matrix ($\metric$ and $\mP$) update intervals $T$, measured on a NVIDIA A100 GPU.
    For the middle row, $T=1$. For the bottom row, $|\minibatch|=128$. Missing points are due to the GPU memory limitation.}
    \label{fig:throughput_memory}
\end{figure*}

\paragraph{Algorithm layer}
This layer defines the high-level behavior of a gradient preconditioning algorithm.
\texttt{PreconditionedGradientMaker} class, which is a child class of \texttt{GradientMaker}, defines the functions common to all \texttt{XXXGradientMaker} classes by the override of the \texttt{forward\_and\_backward} method.
A child of \texttt{PreconditionedGradientMaker} (e.g., \texttt{PsgdGradientMaker}, \texttt{KfacGradientMaker}, and \texttt{SengGradientMaker}) implements following methods:
\begin{enumerate}
    \item \texttt{\textbf{update\_curvature()}}: 
    \texttt{XXXGradientMaker} classes with a \textit{direct} solver (\S\ref{subsec:solver}) implement this method. 
    This method updates a certain representation of the local curvature matrix $\metric_{repr}$ using the information registered by \texttt{setup\_model\_call} and \texttt{setup\_loss\_call/repr} (hereafter, we refer to this as \textit{model-loss information}). 
    If a \textit{global} direct solver (\S\ref{subsec:solver}) is used, the global $\metric_{repr}$ is update by accumulating the calculated local one. 
    
    \item \texttt{\textbf{update\_preconditioner()}}:
    This method updates the preconditioning matrix $\mP$.
    \texttt{XXXGradientMaker} classes with a global iterative solver updates $\mP$ using the current model-loss information while those with a global/local direct solver updates $\mP$ by $\left(\metric_{repr}+\damp\mI\right)^{-1}$.
    In numerical linear algebra, ``solving'' linear equations ($\mA\vx=\vb$) rather than ``inverting'' a matrix ($\mA^{-1}$) is usually preferred in terms of computational cost and accuracy \cite{higham_mixed_2022}. 
    In deep learning, however, it is essential to reduce the frequency of $\mP$ updates (i.e., reuse the stale $\mP$ for some steps) to make gradient preconditioning practical when it is used in training (as observed in \autoref{fig:throughput_memory}), so the inverse matrix needs to be computed explicitly for a direct solver. 
    
    \item \texttt{\textbf{precondition()}}:
    Every \texttt{XXXGradientMaker} class implements this method. 
    This calculates the preconditioned gradient $\mP\vg$ by multiplying $\mP$ to the mini-batch gradient $\vg$ except for a local iterative solver (e.g., Hessian-free), which calculates $\mP\vg$ in an iterative fashion using the current model-loss information only.
\end{enumerate}
\texttt{PreconditionedGradientMaker} class also defines the methods for managing the execution timing of \texttt{update\_curvature} and \texttt{update\_preconditioner} based on the update interval configured via \texttt{XXXGradientConfig}, which is a child class of \texttt{PreconditionedGradientConfig}, and the number of steps so far.
Each of these three methods performs some sort of operations. 
Operations involving the Fisher matrix and operations that require \textit{extensions} to forward/backward passes are delegated to the Matrix layer or Extension layer.

\paragraph{Matrix layer}
The Fisher information matrix $\fim$ (\ref{eq:fim}) and its estimations $\fimnmc$ (\ref{eq:fimnmc}) and $\fimemp$ (\ref{eq:fimemp}) have the same structure:
\begin{align*}
    \avg{\sum_{i=1}^N\nabla h(\nn(\inputs),\targets_i)\nabla h(\nn(\inputs),\targets_i)^\T}\in\R^{\nparams\times\nparams}\,,
\end{align*}
where $\nabla$ is taken w.r.t. $\params$, $N=(\nclasses,n,1)$ and $\targets_i=(\targets'_i,\targets_{\rm mc}^{(i)},\targets)$ for $(\fim,\fimnmc,\fimemp)$, respectively, and we assume $\E_q(\inputs)[\cdot]$ in $\fim$ (\ref{eq:fim}) is replaced with $\avg{\cdot}$.
Therefore, the choice of curvature matrix (\S\ref{subsec:curv}) defines the \textit{inner loop} $\sum$, i.e, the target vector $\targets_i$ and the number of backward passes $N$\footnote{In all cases, forward pass $\nn(\inputs)$ only needs to be evaluated once for each example $\inputs\in\minibatch$.}.
On the other hand, the choice of matrix representation (\S\ref{subsec:repr}) defines the \textit{operations-in-loop}, i.e., how to (approximately) calculate $\nabla h(\cdot)\nabla h(\cdot)^\T$, which is \textit{orthogonal} to the definition of the inner loop and choice of curvature matrix.

Exploiting this relationship, the Matrix layer implements the \texttt{FisherMaker} class which only defines the inner loop for a given Fisher type and loss type (either cross-entropy loss or mean-squared-error loss, only for $\fim$ and $\fimnmc$), and the execution of the operations-in-loop, which are also common to other algorithms without a Fisher matrix, is delegated to the Extension layer.

\paragraph{Extension layer}
The operations for the second-order information (curvature and preconditioning matrices) usually require the batch of \textit{per-example} gradients $\{\nabla\peloss_i\}_{i\in\minibatch}$ rather than the \textit{mini-batch} gradient $\vg=\avg{\nabla\peloss}=\frac{1}{|\minibatch|}\sum_{i=1}^{|\minibatch|}\nabla\peloss_i$.
In PyTorch, we can efficiently compute per-example gradients by utilizing the \texttt{vmap} implemented in functorch\footnote{\url{https://pytorch.org/functorch/stable/}}.
However, a batch of per-example gradients is $|\minibatch|\times\nparams$ in size for a given mini-batch $\minibatch$, and explicitly computing and storing them is not feasible for neural networks with a large $\nparams$.
Fortunately, the hook registration methods of \texttt{torch.nn.Module} (\texttt{.register\_forward\_hook()} and \texttt{.register\_backward\_hook()}\footnote{\url{https://pytorch.org/tutorials/beginner/former_torchies/nnft_tutorial.html}}) allow access to the batch of inputs (or activation) $\{\va_i\}_{i\in\minibatch}$ and gradient w.r.t. outputs (or error) $\{\ve_i\}_{i\in\minibatch}$ of it via \textit{hook functions} during forward and backward passes, respectively, \textit{without any memory overhead}.
These are the ingredients of the per-example gradients --- for a fully-connected layer (\texttt{torch.nn.Linear}), the gradient w.r.t.\ the weight $\nabla\peloss_i=\va_i\otimes\ve_i$, where $\va_i\in\R^{D_{in}}$, $\ve_i\in\R^{D_{out}}$, $D_{in}/D_{out}$ is the input/output dimension, and $\otimes$ is the Kronecker product of vectors --- and we can perform the operations for the second-order information using them in hook functions.

The role of the Extension layer is to \textit{extend} forward and backward passes by registering hook function(s) that performs operations requested from higher layers (the Algorithm and Matrix layers) to each \texttt{torch.nn.Module} with trainable parameters.
The term ``extend'' is inspired by the BackPACK library \cite{dangel_backpack_2020}, which also utilizes the same mechanism to get access to $\{\va_i\}_{i\in\minibatch}$ and $\{\ve_i\}_{i\in\minibatch}$.
The execution of operations are delegated to the Operation layer and the result will be returned after forward and backward passes.

\paragraph{Operation layer}
This layer \textit{schedules} operation executions in response to requests from the Extension layer and \textit{manages} the results .
$\{\va_i\}_{i\in\minibatch}$ and $\{\ve_i\}_{i\in\minibatch}$ are not necessarily ready to be used (e.g., unfolding is required for $\{\va_i\}_{i\in\minibatch}$ in \texttt{torch.nn.Conv2d}), so this layer schedules preprocessing on them before the execution of operations. 
The preprocessing and operations are specific to \texttt{torch.nn.Module}, so the execution is performed in the Module layer.
The same operation (with different arguments) may be performed repeatedly, and this layer is responsible for concatenating or accumulating those results (e.g., $\fim$ requires to accumulate $\nabla h(\cdot)\nabla h(\cdot)^\T$ $\nclasses$ times).

\paragraph{Module layer}
This layer performs preprocessing and operations with the knowledge about its assigned \texttt{torch.nn.Module} such as whether it has the \texttt{bias} parameter or not and the shapes of $\{\va_i\}_{i\in\minibatch}$ and $\{\ve_i\}_{i\in\minibatch}$.

\section{Case Studies with ASDL}
\label{sec:case}
Using ASDL, we compare gradient preconditioning methods for optimization, i.e., adaptive gradient methods (with $\fimempbatch$) and second-order optimization methods (with other $\metric$) with several neural network architectures. 
We target MNIST classification (MLPs) and CIFAR-10 classification (ResNet18, WideResNet28, ViT-tiny, and MLP-Mixer-base) tasks with SGD, AdamW \cite{loshchilov_decoupled_2019}, PSGD (with Kronecker-factored $\mP$), K-BFGS, K-FAC (with $\fimmc$), SENG, and Shampoo (listed in \autoref{tab:representative}).
We use a local solver for K-FAC, i.e., we do not take the running average of mini-batch $\metric$s unlike \citet{martens_optimizing_2015} for comparison purposes.
Following the settings in \citet{yang_sketch-based_2022}, we apply a sketching size of 256 and a truncated SVD of rank 16 for SENG, i.e., per-example activation $\va_i\in\R^{D_{in}\times r}$ and error $\ve_i\in\R^{D_{out}\times r}$ ($i\in\minibatch$, $r=1$ for \texttt{torch.nn.Linear}, and $r=$ output feature map size for \texttt{torch.nn.Conv2d}) are replaced with matrices of size $\min(D_{in},256)\times\min(r,16)$ and $\min(D_{out},256)\times\min(r,16)$, respectively, before calculating the information of $\metric$.

\subsection{Throughput and memory}
\autoref{fig:throughput_memory} shows the peak memory consumption and throughput (image/s) compared to SGD in training several neural networks on MNIST and CIFAR-10 classification.
SMW formula-based methods such as SENG achieve relatively low memory and high throughput when $|\minibatch|$ (\texttt{Batch size}) is small (e.g., $32$), however, as they involve a $\gO(|\minibatch|^3)$ computational cost and a $\gO(|\minibatch|^2)$ memory cost, they scale badly with $|\minibatch|$. In addition, they are often infeasible for sequencing models such as ViT and MLP-Mixer because $|\minibatch|$ corresponds to the number of tokens, making them particularly compute and memory intensive.
For the other methods, increasing $|\minibatch|$ leads to smaller memory ratio and higher throughput ratio compared to SGD of the same $|\minibatch|$.
This is because the main computational and memory overhead in these methods, i.e., operations for $\metric$ and $\mP$, which are often independent of $|\minibatch|$, become relatively smaller than the costs of forward and backward passes as $|\minibatch|$ grows. 
As Shampoo performs an eigenvalue decomposition much heavier than a matrix inversion, it is relatively slow especially in large networks. 
Still, it benefits most from increasing $|\minibatch|$ as it has no overhead depending on $|\minibatch|$.

With the given \textit{matrix update interval} (\texttt{Interval}) $T>1$, \texttt{update\_curvature()} for calculating $\metric$ and \texttt{update\_preconditioner()} for calculating $\mP$ (discussed in \autoref{subsec:structure}) are called only every $T$ training steps and the stale preconditioning matrix will be reused for $T-1$ steps, which significantly improves the throughput of every method (the memory consumption is not affected).

\begin{table*}
    \centering
    \caption{
    \textbf{The test accuracy} for models achieving the best validation accuracy. 
    For each task, the best accuracy is \textbf{bolded}.
    ``w'': width.
    For ResNet18, the results with 20 and 100 epochs are shown (the number of epochs is fixed for the others).
    SENG consumes lots of memory and is infeasible with MLP-Mixer-base.
    The training settings are described in \autoref{app:setting}.
    }
    \resizebox{\textwidth}{!}{
        \begin{tabular}{lcccccccc}
        \toprule
        \multirow{2}[2]{*}{{Method}} & \multicolumn{3}{c}{MNIST} & \multicolumn{4}{c}{CIFAR-10} \\
        \cmidrule(lr){2-4}\cmidrule(lr){5-8}
        & MLP (w=128) & MLP (w=512) & MLP (w=2048) & ResNet18 & WideResNet28 & ViT-tiny & MLP-Mixer-base \\
        \midrule
        \midrule
        SGD & \textbf{98.9} & 99.1 & \textbf{99.2} & 91.2 / 95.7 & 96.7 & 97.8 & 97.2 \\
        AdamW & 98.7 & 99.0 & 99.1 & 89.9 / 94.8 & 96.0 & 97.9 & \textbf{97.7} \\
        \midrule 
        PSGD (KF) & \textbf{98.9} & 99.1 & \textbf{99.2} & 93.3 / \textbf{96.2} & 96.6 & \textbf{98.0} & 97.5 \\
        K-BFGS & 98.7 & 98.9 & 99.0 & 91.4 / 95.7 & 96.5 & 97.7 & 97.5 \\
        K-FAC (1mc) & 98.8 & \textbf{99.2} & \textbf{99.2} & \textbf{93.6} / 96.1 & \textbf{96.9} & 97.4 & \textbf{97.7} \\
        SENG & 98.8 & 99.0 & 99.1 & 91.6 / 95.8 & 96.6 & 97.7 & - \\
        Shampoo & 98.8 & 99.1 & \textbf{99.2} & 92.5 / 96.1 & \textbf{96.9} & \textbf{98.0} & 97.4 \\
        \bottomrule
        \end{tabular}
    }
    \label{tab:testacc}
\end{table*}

\subsection{Training results and parameter sensitivity}
\begin{figure*}[t]
    \centering
    \includegraphics[width=\textwidth]{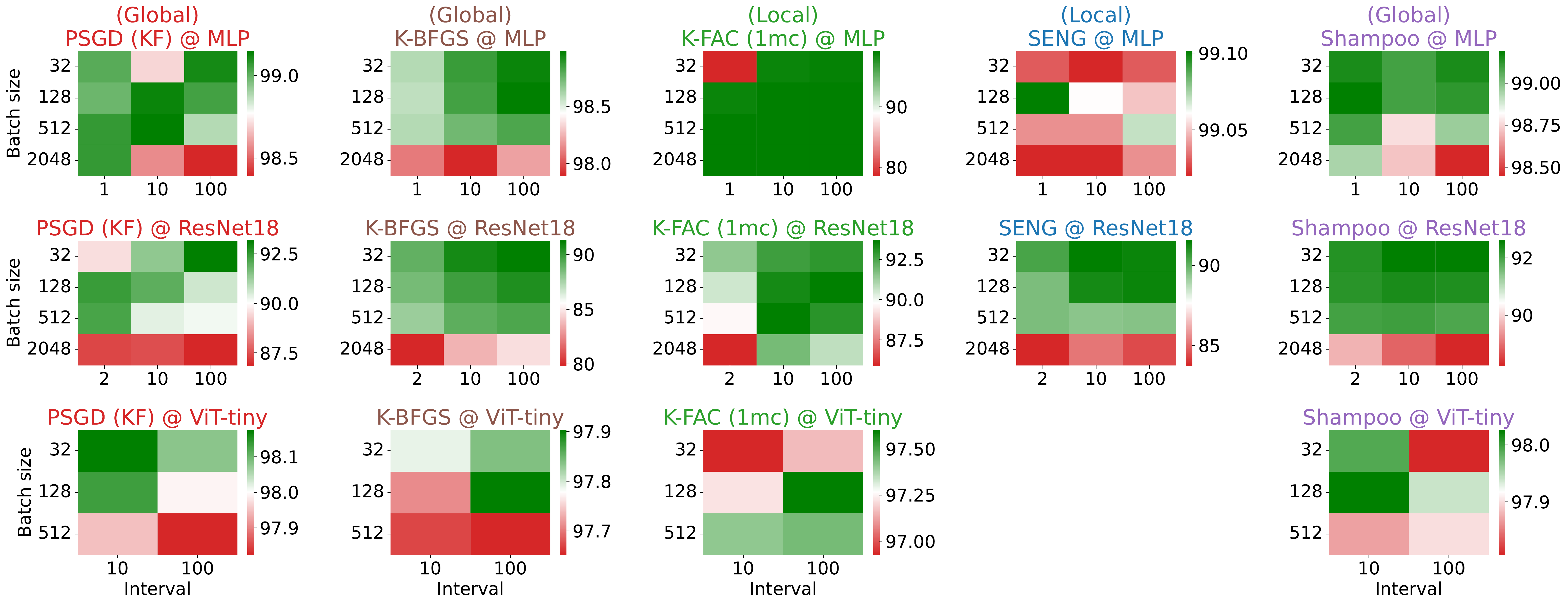}
    \caption{
    Sensitivity of the mini-batch size $|\minibatch|$ and matrix update interval to the test accuracy (the best value among different learning rates for each pair is shown).
    The type of the solver (\S\ref{subsec:solver}) (``Global'' or ``Local'') is indicated at the top of each column.
    For SENG at ViT-tiny, the plot is not shown because it is not feasible with large mini-batch sizes and only B=32 results are available.
    }
    \label{fig:bs_interval_acc}
\end{figure*}

\begin{figure*}[h!]
    \centering
    \includegraphics[width=.8\textwidth]{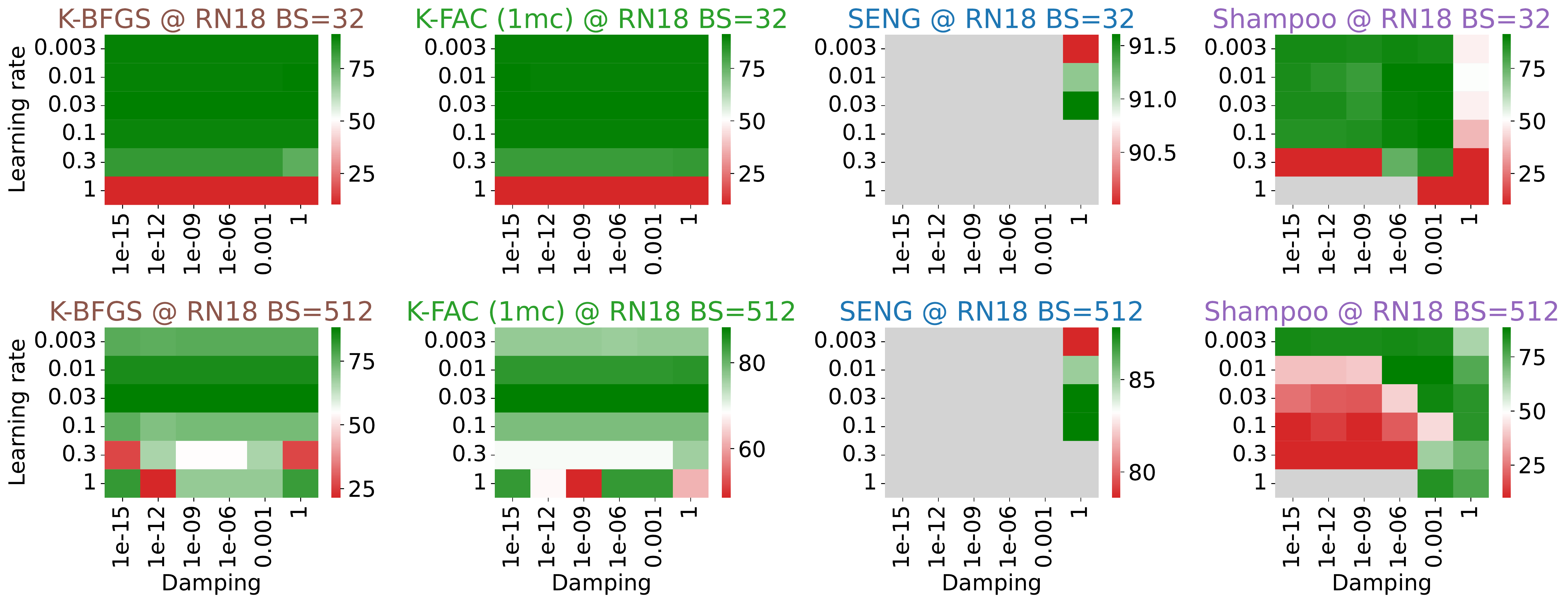}
    \caption{Sensitivity of the learning rate and damping value $\damp$ to the test accuracy in training ResNet18 (RN18) on CIFAR-10 with mini-batch size $|\minibatch|$ of 32 and 512.
    PSGD does not take a damping value, so it is excluded in this comparison.
    The gray boxes indicate that the training with corresponding learning and damping value diverged.}
    \label{fig:lr_damp_acc_rn18}
\end{figure*}

\autoref{tab:testacc} summarizes the training results. 
The best test accuracy for each task is achieved by one of the gradient preconditioning methods, but the best performing method depends on the task.
\autoref{fig:bs_interval_acc} summarizes the test accuracy of MLP ($\mathrm{width}=512$) models on MNIST or ResNet18/ViT-tiny models on CIFAR-10 classification trained for $20$ epochs with different mini-batch sizes $|\minibatch|$ and matrix update intervals $T$.
Methods with a global solver (\S\ref{subsec:solver}), i.e., PSGD, K-BFGS, and Shampoo, tend to achieve a lower accuracy with larger $|\minibatch|$ and $T$.
One possible explanation is that the preconditioning matrix $\mP$ is \textit{immature} because the number of updates of $\mP$ per epoch becomes smaller when $|\minibatch|$ and $T$ are larger.
On the other hand, K-FAC with a local solver, which only includes information on one $\minibatch$ in $\mP$, tend to achieve \textit{higher} accuracy with larger $|\minibatch|$ and $T$.
One possible reason for the better accuracy with a larger $T$ is that fitting to a particular mini-batch (which does not represent the data distribution well) with a \textit{too accurate} descent direction (given by $\mP\vg$) is detrimental to the overall training loss and test performance.
The other ``local'' method, SENG, is very sensitive to the hyperparameters (i.e., learning rate and damping value $\damp$), as seen in \autoref{fig:lr_damp_acc_rn18} (for ResNet18), and does not share the same characteristics as K-FAC.

\section{Related Work}
The studies most relevant to this study are the BackPACK \cite{dangel_backpack_2020} and NNGeometry \cite{george_nngeometry_2021}, which are also extension libraries of PyTorch for calculating the Kronecker-factored or diagonal second-order matrices ($\ggn$, $
\fim$, $\fimnmc$, and $\fimemp$).
PyHessian \cite{yao_pyhessian_2020} is also a PyTorch-based library which calculates $\hess$ and estimates its eigenvalues. 
Compared to them, ASDL offers a more comprehensive selection of curvature and matrix representation combinations. 
In addition, while they only focus on matrix calculations, ASDL also facilitates a flexible matrix \textit{utilization} via various implementations and a unified interface for gradient preconditioning.

\section{Discussion and Conclusion}
\paragraph{Future work}
The current version of ASDL does not support distributed and mixed-precision training, where time and numerical stability bottlenecks change \citep{ueno_rich_2020,anil_scalable_2021}.
Extending this work to these training settings is an important future direction, and the unified interface (\S\ref{subsec:interface}, \S\ref{subsec:versatile}) and hierarchical abstraction structure (\S\ref{subsec:structure}) in ASDL facilitate such extensions.

\paragraph{Conclusion}
Using ASDL, we observe that no gradient preconditioning method is always superior (in computing performance, prediction accuracy, and feasibility) to another --- it is critical to switch and compare methods flexibly.
In addition, since gradient preconditioning is particularly complex to implement in deep learning training pipelines, it is undesirable to duplicate implementation, debugging, and testing efforts among researchers.
We believe ASDL and its unified interface will facilitate fair and structured comparisons and quick adaptations of gradient preconditioning methods in deep learning of wide domains and applications.


\bibliography{references.bib}
\bibliographystyle{mlsys2023}

\appendix

\newpage
\begin{table*}[]
    \centering
    \caption{
    \textbf{Gradient preconditioning methods in deep learning}. 
    ``\textbf{KF-io}'': input-output Kronecker-factored. 
    ``\textbf{KF-dim}'': dimension-wise Kronecker-factored. 
    ``\textbf{RR}'': rank reduction. 
    ``\textbf{SMW}'': Sherman-Morrison-Woodbury formula.
    ``\textbf{L}'': local
    ``\textbf{G}'': global
    ``\textbf{iter}'': iterative.
    ``\textbf{NN ind.}'': how to calculate $\mP\vg$ is independent of the neural network architecture.
    If the matrix $\metric$ is ``\textit{full}'' granularity, it can be applied to any granularity (e.g., PSGD (KF), TONGA (unit) introduced by the authors), but some methods require additional derivation, computation and memory costs.
    }
    \resizebox{\textwidth}{!}{
        \begin{tabular}{lllllllcc}
        \toprule
        \multirow{2}[2]{*}{{Method}} & \multicolumn{2}{c}{\textbf{Curvature} matrix $\metric$ (\S\ref{subsec:curv})} & \multicolumn{2}{c}{\textbf{Representation} of $\metric$ (\S\ref{subsec:repr})} & \multicolumn{2}{c}{\textbf{Solver} for $\mP\vg\approx\metric^{-1}\vg$ (\S\ref{subsec:solver})} & \multirow{2}[2]{*}{{NN ind.}} \\
        \cmidrule(lr){2-3}\cmidrule(lr){4-5}\cmidrule(lr){6-7}
        & type & matrix & granularity & format & type & key operations \\
        \midrule
        LiSSA \citep{agarwal_second-order_2017} & \sharpness & $\hess$ & full & dense & G iter & Neumann series & \checkmark \\
        PSGD \citep{li_preconditioned_2018} & \sharpness & $\hessabs$ & full & dense & G iter & triangular solve \& SGD & \checkmark \\
        \scriptsize{Neumann optimizer} \citep{krishnan_neumann_2017} & \sharpness & $\hess$ & full & matrix-free & L iter & Neumann series & \checkmark \\
        Hessian-free \citep{martens_deep_2010} & \sharpness & $\hess,\ggn$ & full & matrix-free & L iter& conjugate gradient & \checkmark \\
        {KSD} \citep{vinyals_krylov_2011} & \sharpness  & $\hess,\ggn$ & full & matrix-free & L iter & Krylov subspace method & \checkmark \\
        L-BFGS \citep{liu_limited_1989} & \sharpness & $\hessbfgs$ & full & matrix-free & G iter & approx. BFGS & \checkmark \\
        SMW-GN \citep{ren_efficient_2019} & \sharpness & $\ggn$ & full & Gram, RR & L direct & SMW inverse & \xmark \\
        SMW-NG \citep{ren_efficient_2019} & \gsecm & $\fimemp$ & full  & Gram, RR & L direct & SMW inverse & \xmark \\
        TONGA \citep{roux_topmoumoute_2008} & \gsecm & $\fimemp$ & full & Gram, RR & G direct & SMW solve \& eigendecomp. & \checkmark \\
        M-FAC \citep{frantar_efficient_2021} & \gsecm & $\fimempbatch$ & full & Gram, RR & G direct & SMW solve & \checkmark \\
        GGT \citep{agarwal_efcient_2019} & \gsecm & $\sqrtfimempbatch$ & full & Gram, RR & G direct & SMW solve & \checkmark \\
        \scriptsize{FANG} \citep{grosse_scaling_2015} & \gcov & $\fimnmc$ & full & sparse & L/G direct & incomplete Cholesky & \checkmark \\
        \midrule
        PSGD (KF) \citep{li_preconditioned_2018} & \sharpness & $\hessabs$ & layer & KF-io & G iter & triangular solve \& SGD & \xmark \\
        K-BFGS \citep{goldfarb_practical_2021} & \sharpness & $\hessbfgs$ & layer & KF-io & G iter & BFGS & \xmark \\
        K-FAC \citep{martens_optimizing_2015} & \gcovsecm & $\fimnmc,\fimemp$ & layer & KF-io & L/G direct & Cholesky inverse & \xmark \\
        KFLR \citep{botev_practical_2017} & \gcov & $\fim$ & layer & KF-io & L/G direct & Cholesky inverse & \xmark \\
        KFRA \citep{botev_practical_2017} & \gcovsecm & $\fimnmc,\fimemp$ & layer & KF-io & L/G direct & Cholesky inverse \& recursion & \xmark \\
        EKFAC \citep{george_fast_2018} & \gcovsecm & $\fimemp$ & layer & KF-io & L/G direct & eigendecomp. (or SVD) & \xmark \\
        SKFAC \citep{tang_skfac_2021} & \gcovsecm & $\fimmc,\fimemp$ & layer & KF-io, RR & L direct & SMW inverse \& reduction & \xmark \\
        SENG \citep{yang_sketchy_2021} & \gsecm & $\fimemp$ & layer & Gram, RR & L/G direct & SMW inverse \& sketching & \xmark \\
        TNT \citep{ren_tensor_2021} & \gcovsecm & $\fimnmc,\fimemp$ & layer & KF-dim & L direct & Cholesky inverse & \checkmark \\
        Shampoo \citep{gupta_shampoo_2018} & \gsecm & $\sqrtfimempbatch$ & layer & KF-dim & G direct & eigendecomp. & \checkmark \\
        \midrule
        \small{unit-wise NG} \citep{ollivier_riemannian_2015} & \gcovsecm & $\fimnmc,\fimemp$ & unit & dense & L/G direct & Cholesky inverse & \xmark \\
        \scriptsize{TONGA (unit)} \citep{roux_topmoumoute_2008} & \gsecm & $\fimemp$ & unit & Gram, RR & G direct & SMW solve \& eigendecomp. & \xmark \\
        \midrule
        AdaHessian \citep{yao_adahessian_2020} & \sharpness & $\hessabs$ & element & dense & G direct & element-wise division & \checkmark \\
        SFN \citep{dauphin_identifying_2014} & \sharpness & $\hessabs$ & element & dense & L/G direct & element-wise division & \checkmark \\
        \scriptsize{Equilibrated SGD} \citep{dauphin_equilibrated_2015} & \sharpness & $\hessabs$ & element & dense & L/G direct & element-wise division & \checkmark \\
        AdaGrad \citep{duchi_adaptive_2011} & \gsecm & $\sqrtfimempbatch$ & element & dense & G direct & element-wise division & \checkmark \\
        Adam \citep{kingma_adam_2015} & \gsecm & $\sqrtfimempbatch$ & element & dense & G direct & element-wise division & \checkmark \\
        \bottomrule
        \end{tabular}
    }
    \label{tab:prec_full}
\end{table*}

\section{Experimental settings}
\label{app:setting}

We split the training set of MNIST (60,000 images) into 49,152 and 10,848 images for training and validation, respectively, and evaluate the test accuracy using the testing set (10,000 images).
Similarly, we split the training set of CIFAR-10 (50,000 images) into 45,056 and 4,944 images for training and validation, respectively, and evaluate the test accuracy using the testing set (10,000 images).
For each task, we tune the mini-batch size, initial learning rate, number of epochs, matrix update interval (for PSGD, K-BFGS, K-FAC, SENG, and Shampoo), and damping $\damp$ (for K-BFGS, K-FAC, SENG, and Shampoo) using a grid search.
The learning rate is schedule by the cosine annealing decay so that it becomes 0 at the end of training (i.e., the number of epochs affects the decaying speed of learning rate). 
We apply gradient clipping with the maximum norm of 1.
For each task and method, we report the test accuracy of the model checkpoint (in every epoch) achieving the best validation accuracy in \autoref{tab:testacc}.
As a baseline, we also train models with SGD with momentum of 0.9 and AdamW with the default parameters in PyTorch\footnote{\url{https://pytorch.org/docs/stable/generated/torch.optim.AdamW.html}} except for the learning rate and weight decay.

\subsection{MLP on MNIST}
We train three-layer multilayer perceptron (MLP) models with a width of 128, 512, or 2048.

\begin{itemize}
    \item Mini-batch size : \{32,128,512,2048\}
    \item Initial learning rate : \{3e-1,1e-1,3e-2,1e-2,3e-3,1e-3\}
    \item Number of epochs : 20
    \item Matrix update interval (PSGD, K-BFGS, K-FAC, SENG, and Shampoo) : \{1,10,100\}
    \item Damping $\damp$ (for SENG, K-FAC, and Shampoo) : 1e-3
    \item Damping $\damp$ (for K-BFGS) : 1e-6
    \item Global norm of gradient clipping : 10
\end{itemize}
We use a weight decay of 5e-4 and apply no data augmentation.

\subsection{ResNet18 and WideResNet on CIFAR-10}
We use WideResNet with a depth of 28.
We use the existing implementation\footnote{\url{https://github.com/uoguelph-mlrg/Cutout}} for defining these architectures.
For training WideResNet we adopt dropout(droprate=0.3).


\begin{itemize}
    \item Mini-batch size : \{32,128,512,2048\}
    \item Initial learning rate : \{3e-1,1e-1,3e-2,1e-2,3e-3,1e-3\}
    \item Number of epochs : 100
    \item Matrix update interval (for PSGD, K-BFGS, K-FAC, SENG, and Shampoo) : \{10,100\}
    \item Damping $\damp$ (for K-FAC, SENG, and Shampoo) : 1e-3
    \item Damping $\damp$ (for K-BFGS) : 1e-6
    \item Global norm of gradient clipping : 10
\end{itemize}

We use a weight decay of 5e-4.
We apply RandomCrop, RandomHorizontalFlip and Cutout as data augmentation.

\subsection{ViT-tiny and MLP-Mixer-base on CIFAR-10}
We fine-tune ViT-T/16 and Mixer-B/16 models pretrained on ImageNet-1K.


\begin{itemize}
    \item Mini-batch size : \{32,128,512\}
    \item Initial learning rates : \{3e-1,1e-1,3e-2,1e-2,3e-3,1e-3\}
    \item Number of epochs : 20
    \item Matrix update interval (for PSGD, K-BFGS, K-FAC, SENG, and Shampoo) : \{10,100\}
    \item Damping $\damp$ (for K-FAC, SENG, and Shampoo) : 1e-3
    \item Damping $\damp$ (for K-BFGS) : 1e-6
    \item Global norm of gradient clipping : 10
\end{itemize}

We use a weight decay of 1e-4.
We apply RandomCrop, RandomHorizontalFlip and Cutout as data augmentation.

\end{document}